\documentclass[11pt]{article}

\usepackage[preprint]{acl}

\usepackage{times}
\usepackage{latexsym}
\usepackage{booktabs,tabularx,array}
\usepackage{placeins}     % \FloatBarrier
\usepackage[T1]{fontenc}
\usepackage[utf8]{inputenc}

\usepackage{stfloats}     % better two-col float handling
\usepackage{adjustbox}    % \begin{adjustbox}{max width=...}

\usepackage{acl}
\usepackage{microtype}
\usepackage[most]{tcolorbox}
\usepackage{enumitem}     % compact lists
\usepackage{ragged2e}     % \RaggedRight (better than \raggedright for code-like text)
\tcbset{
  appbox/.style={
    colback=gray!7, colframe=black!30, boxrule=0.6pt, arc=5pt,
    left=2mm, right=2mm, top=1.5mm, bottom=1.5mm, width=\linewidth
  }
}
\usepackage[font=small,labelfont=bf]{caption}
\newlist{tightenum}{enumerate}{1}
\setlist[tightenum]{nosep,leftmargin=*,label=\arabic*.}

\usepackage{inconsolata}
\usepackage{multirow}

\usepackage{graphicx}
\usepackage{amsmath}
\DeclareRobustCommand{\cvScoreSymb}{%
  \ensuremath{\widehat{\rho}_{\ell,p}^{\mathrm{CV}}}%
}

\usepackage{tcolorbox}
\tcbuselibrary{skins, breakable}

\newtcbox{\alphabadge}{nobeforeafter, tcbox raise base,
  colback=black!5, colframe=black!30, boxrule=0.3pt, arc=2pt,
  left=2pt, right=2pt, top=0.6pt, bottom=0.6pt}

\newtcolorbox{questionbox}{
  enhanced, colback=black!02, colframe=black!08,
  boxrule=0.3pt, arc=6pt, left=6pt, right=6pt, top=5pt, bottom=5pt
}

\usepackage{varwidth}
\definecolor{badhl}{RGB}{255,230,230}
\definecolor{goodhl}{RGB}{227,246,231}
\definecolor{codehl}{RGB}{232,240,254}
\newcommand{\hlbad}[1]{\colorbox{badhl}{#1}}
\newcommand{\hlgood}[1]{\colorbox{goodhl}{#1}}
\newcommand{\hlcode}[1]{\colorbox{codehl}{\ttfamily #1}}

\newtcolorbox{diffbox}[1][]{
  enhanced, breakable, colback=white, colframe=black!10, 
  boxrule=0.4pt, arc=6pt, left=6pt, right=6pt, top=5pt, bottom=5pt,
  fonttitle=\bfseries, title=#1
}

\title{LLMs Encode How Difficult Problems Are}

\author{
  William Lugoloobi\thanks{Corresponding author.} \quad Chris Russell \\
  University of Oxford \\
  \texttt{\{first.last\}@oii.ox.ac.uk}
}

\begin{document}
\maketitle
% \begin{abstract}
% We ask whether large language models encode a notion of problem difficulty that aligns with humans, and whether this representation tracks generalisation during RL post-training. We train linear probes over layers and token positions and evaluate on math and coding subsets of Easy2HardBench. Across 60 models, human-labelled difficulty is strongly linearly decodable (e.g., AMC best $\rho\approx$ 0.88) and exhibits clear size scaling, whereas LLM-derived difficulty is weaker and scales poorly. During GRPO on Qwen2.5-Math-1.5B, the human-difficulty probe strengthens and positively co-varies with test Pass@1 beyond training step, while the LLM-difficulty probe degrades and negatively co-varies. These results suggest that human annotations supply a stable difficulty direction that RL amplifies, while automated difficulty from model performance becomes misaligned  as models improve. We release probe code and evaluation scripts to facilitate replication. 
% \end{abstract}
\begin{abstract}
Large language models exhibit a puzzling inconsistency: they solve complex problems yet frequently fail on seemingly simpler ones. We investigate whether LLMs internally encode problem difficulty in a way that aligns with human judgment, and whether this representation tracks generalization during reinforcement learning post-training. We train linear probes across layers and token positions on 60 models, evaluating on mathematical and coding subsets of Easy2HardBench. We find that human-labeled difficulty is strongly linearly decodable (AMC: $\rho \approx 0.88$) and exhibits clear model-size scaling, whereas LLM-derived difficulty is substantially weaker and scales poorly. Steering along the difficulty direction reveals that pushing models toward "easier" representations reduces hallucination and improves accuracy. During GRPO training on Qwen2.5-Math-1.5B, the human-difficulty probe strengthens and positively correlates with test accuracy across training steps, while the LLM-difficulty probe degrades and negatively correlates with performance. These results suggest that human annotations provide a stable difficulty signal that RL amplifies, while automated difficulty estimates derived from model performance become misaligned precisely as models improve. We release probe code and evaluation scripts to facilitate replication. 
\end{abstract}

\section{Introduction}

\begin{figure}[t]
\centering

\begin{questionbox}
\footnotesize
\textbf{Question.}\quad Let $(a_1,\dots,a_n)$ be positive reals with
$\sum a_i=96$, $\sum a_i^2=144$, $\sum a_i^3=216$. \emph{Find the sum of all possible values of $n$.}
\end{questionbox}
% \vspace{3pt}
% Steered (good, left accent)
\begin{diffbox}[Steered \alphabadge{$\alpha=-3$}, coltitle=black]
\footnotesize
\begin{varwidth}{\columnwidth}\raggedright
\textbf{Response:}\
“Let’s solve step-by-step using \hlcode{Python}/\hlcode{SymPy}.
Given $\sum a_i=96$, $\sum a_i^2=144$, $\sum a_i^3=216$, we solve for admissible $n$ via
Cauchy–Schwarz / power-mean constraints… \ldots\ \hlgood{64}.
(Use \hlcode{import sympy as sp}, \hlcode{factor()}, \hlcode{solve()} to verify.)”
\\[3pt]
\textbf{Parsed answer:}\ \hlgood{64}
\textbf{Response length:} \hlgood{1298}\\
\end{varwidth}
\end{diffbox}
% \vspace{3pt}
% Baseline (and +3) joined, with subtle red accent
\begin{diffbox}[Baseline \alphabadge{$\alpha=0,\;+\!3$}, coltitle=black]
\footnotesize
\begin{varwidth}{\columnwidth}\raggedright
\textbf{Response:}\
\hlbad{legitim} \hlbad{legitim} \hlbad{legitim} \hlbad{legitim} \hlbad{legitim} \hlbad{legitim}
\hlbad{legitim} \hlbad{legitim} \hlbad{legitim} \hlbad{legitim} \ldots\\[3pt]
\textbf{Parsed answer:}\ \hlbad{none}
\textbf{Response length:} \hlbad{3000}\\
\end{varwidth}
\end{diffbox}
% \vspace{4pt}
\caption{\textbf{Steering towards “easy” reduces hallucination-like repetition in Qwen2.5-Math-1.5B.}
With $\alpha=-3$, the model switches from repetitive output (baseline / $+\!3$) to concise, code-assisted reasoning and a parsed correct answer (\(64\)).}
\label{fig:qual-steer-card}
\end{figure}

Large language models (LLMs) have demonstrated remarkable performance on complex tasks. However, they exhibit a fundamental inconsistency: they solve challenging problems while frequently failing at seemingly simpler ones \cite{yang_can_2024, williams_easy_2024, wang_is_2025}. This inconsistency manifests in various forms, from failures in basic logical reasoning to difficulties with symmetric relationships. As LLMs are deployed in agentic and high-risk applications, understanding and mitigating this inconsistency becomes increasingly critical.

A natural question thus arises: do LLMs internally represent problem difficulty in a way that aligns with human judgment? While LLMs fail when explicitly prompted to estimate problem difficulty \cite{ding_easy2hard-bench_2024} and struggle to calibrate their confidence on challenging questions \cite{kadavath_language_2022}, this does not mean that difficulty is absent from their internal representations. Recent work has demonstrated that high-level concepts such as truthfulness and sentiment are linearly encoded in LLM activations \cite{mikolov_linguistic_2013, park_linear_2024}, suggesting that difficulty might similarly exist as a recoverable direction in the model's representation space—even if models do not articulate it when prompted.

In this work, we investigate whether LLMs encode a notion of problem difficulty that aligns with human judgment, examine the effects of steering along this direction, and track how this representation evolves during Reinforcement Learning from Verifiable Rewards (RLVR) for mathematical reasoning. We train linear probes across 60 models and evaluate their difficulty estimation performance on mathematical and coding subsets from Easy2HardBench \cite{ding_easy2hard-bench_2024}. 

Our key findings are:
\begin{itemize}
    \item Human-derived difficulty ratings are strongly and linearly decodable from model activations, while LLM-derived difficulty ratings show substantially weaker encoding.
    \item Steering along the difficulty direction reveals that pushing models toward "easier" representations reduces output length and improves accuracy by preventing hallucination.
    \item During GRPO training, human-difficulty representations strengthen alongside model capabilities, whereas LLM-difficulty representations degrade—suggesting that human annotations capture a stable signal that RL amplifies, while automated difficulty estimates become increasingly misaligned as models improve.
\end{itemize}

\section{Background}

\subsection{Linear Probes}

Substantial previous work has shown that certain high-level features such as truthfulness and political perspectives, may be linearly represented in the activations of an LLM \citep{mikolov_linguistic_2013, park_linear_2024, kim_linear_2025}. Linear probes are simple models trained on activations extracted from an LLM at a given timestep $t$ to predict a "ground truth" label $y$. Some existing work advocates for probing the outputs of individual attention heads  and fitting a separate probe to each head \cite{kim_linear_2025}. A contrasting approach is to probe the outputs of the concatenated attention outputs from the attention block at each layer of the model. \citep{kim_linear_2025} %Remove unless you have citations

\subsection{Scaling Laws}
Previous research on scaling laws has demonstrated that architectural and training regime differences lead to divergent behaviours for similar model sizes across different model families \citep{kaplan_scaling_2020}. Consequently, we conduct our probing study across five major model families—Qwen2.5, Qwen3, LLaMA 3.1, DeepSeek \cite{team_qwen25_2024,team_qwen3_2025, llama_team_llama_2024,deepseek-ai_deepseek-r1_2025}—to observe whether difficulty is linearly encoded among them.

\subsection{Difficulty estimation}
Traditional difficulty estimation relies on domain-specific characteristics like language similarity in linguistic reasoning \citep{beeching_open_2023}, equation complexity in mathematical problems \citep{yang_can_2024}, or human annotations \citep{rein_gpqa_2023}.

More recently, \citet{ding_easy2hard-bench_2024} introduced Easy2Hard-Bench, a dataset that aggregates multiple benchmarks spanning various difficulty levels. The dataset provides standardized, human-grounded difficulty labels across six subsets: E2H-AMC, E2H-Codeforces, E2H-Lichess, E2H-GSM8K, E2H-Winogrande, and E2H-ARC. These difficulty ratings are derived from two primary sources: human performance statistics from competitive leaderboards (for E2H-AMC, E2H-Codeforces, and E2H-Lichess) and LLM performance data from the Open LLM Leaderboard \citep{beeching_open_2023} (for E2H-GSM8K, E2H-ARC, and E2H-Winogrande). For the AMC subset specifically, difficulty is estimated using Item Response Theory (IRT) models fitted on multiple attempts by humans.

\subsection{RLVR}
Popular LLM's such as OpenAI-o1 \citep{openai_learning_2024}, DeepSeek-R1 \citep{deepseek-ai_deepseek-r1_2025}, and Kimi-1.5 \citep{team_kimi_2025} have used Reinforcement Learning (RL) as a post-training approach to enhance the reasoning capabilities of models and consequently improve their performance on complex mathematical tasks. \textit{Reinforcement Learning with Verifiable Rewards} (RLVR) \citep{deepseek-ai_deepseek-r1_2025, gao_designing_2024, wang_reinforcement_2025, liu_understanding_2025} is a key method central to these advances in which RL is incorporated in an LLM with a rule-based outcome reward, such as a binary reward for the accuracy of a model’s final answer to a math problem or following a template of expected answers. RLVR is commonly optimised with policy-gradient methods such as PPO and GRPO.

Recent work therefore targets \emph{selective sampling} by difficulty variability. \citet{wang_reinforcement_2025} demonstrate that performance comparable to or better than GRPO with full-dataset training can be achieved with one or two carefully selected problems, chosen using a \emph{historical variance} score that tracks fluctuations in the model’s success rate across steps. Complementarily, \citet{foster_learning_2025} propose LILO, which measures \emph{within-step} success variability to surface challenging items online. Both strategies rely on observing variance—either across steps (RLVR) or within a step (LILO)—and typically require an initial warm-up or logging pass to identify high-variance problems.

\section{Experiments}

\subsection{Probing for Difficulty}\label{subsec:Probe_Setup}
\paragraph{Data} We construct our probing datasets using two mathematical subsets from Easy2Hard-Bench~\citep{ding_easy2hard-bench_2024}: E2H-AMC and E2H-GSM8K. Each problem has a difficulty score derived from Item Response Theory (IRT) modeling, where higher scores indicate greater difficulty. The key difference between the subsets is that E2H-AMC uses human success rates to calibrate the difficulty, while E2H-GSM8K uses LLM success rates. For each subset, we randomly sample 500 questions from the original training set to create our probing dataset $\mathcal{D}_{\text{train}}$. 

\paragraph{Models}
We extract activations from eighteen model families across 60 variants: DeepSeek, Qwen-2.5, Qwen-Math, Llama, Qwen-3, FineMath, OpenMath, Qwen-Coder. To assess whether domain-specific fine-tuning affects difficulty representations, we include math-specialized variants—FineMath and OpenMath (Llama-3.1 based) and Qwen-Math (Qwen-2.5 based)—for the AMC and GSM8K subsets. We also include reasoning-focused models (DeepSeek and Qwen-3) to examine whether enhanced reasoning capabilities yield better difficulty representations. For the coding dataset, we add two code-specialized Qwen-Coder variants (OpenHands and SWE-Agent) following the same rationale. A complete list of models is provided in Appendix~\ref{app:tab:models_list}.

\paragraph{Linear Probe Setup}
For each problem, we apply the model's default chat template and extract activations from all tokens following the final prompt token (post-instruction tokens). At each post-instruction token position, we extract activations from every model layer $l$.

For each problem, we format the input with the model’s chat template and collect activations from every token after the final prompt token (“post-instruction” tokens). Therefore, at each such position, $p$, we extract the hidden states from every layer $l$.

\paragraph{Evaluation.} We evaluate all linear probes using 5-fold cross-validation on the training split $\mathcal{D}_{\text{train}}$. For each fold, we train a probe on 80\% of the data and evaluate on the remaining 20\%. The metric is Spearman rank correlation between predicted and ground-truth difficulty. For each (layer, position) probe we report the mean across folds, denoted \cvScoreSymb.

To assess the effect of model size on performance, we fit a model's best \cvScoreSymb to the following power law:
$$1 - \text{\cvScoreSymb} = C \cdot N^{-\alpha}$$
where $N$ is the number of model parameters, $\alpha$ is the scaling exponent (with $\alpha > 0$ indicating performance improves with model size), and $C$ is a normalization constant. This formulation models the gap to perfect performance, allowing performance to asymptotically approach 1.0 as model size increases. We perform the fit in log-log space by regressing $\log(1 - \text{perf})$ against $\log(N)$, ensuring numerical stability through epsilon clipping. We report $R_{log}^2$ and $\alpha$, which measure the proportion of variance in the log-transformed performance gap explained by log model size.

Assuming that we observe strong performance of our linear probes, we can assume that a linear "easy-hard" axis exists within LLM representations. Following \citet{li_inference-time_2024}, we can then use the fitted regression coefficients from our trained probes to steer generation during inference and observe how this affects task performance. In our case, we define a steering coefficient grid ranging from $\alpha \in [-3, +3]$, and our metric of interest will be Pass@1.

% We explore steering with difficulty vectors on Qwen2.5-Math-1.5B's performance on the MATH500 benchmark \cite{hendrycks_measuring_2021}, defining a steering coefficient grid ranging from $\alpha \in [-3, +3]$.

\subsubsection{Results}
\paragraph{Human Difficulty is Linearly Encoded in LLM Activations}
% \begin{table}[t]
%   \centering
%   \footnotesize
%   \setlength{\tabcolsep}{5pt}
%   \renewcommand{\arraystretch}{1.15}
%   \begin{tabularx}{\columnwidth}{l >{\raggedright\arraybackslash}X c c c}
%     \toprule
%     \textbf{Dataset} & \textbf{Model} & \textbf{Test Perf} & \textbf{Layer} & \textbf{Pos.} \\
%     \midrule
%     \midrule
%     \textbf{AMC}        & \textbf{DS-R1-LLaMA-70B} & \textbf{0.88} & \textbf{38} & \textbf{-1} \\
%     Codeforces & LLaMA-3.1-70B-Inst.           & 0.81 & 42 & -1 \\
%     GSM8K      & LLaMA-3.1-70B-Inst.           & 0.58 & 35 & -1 \\
%     \bottomrule
%   \end{tabularx}
%   \caption{Top-performing difficulty probes for each dataset. The Test Performance is the average Spearman Rank correlation over 5 folds (\cvScoreSymb). Best results are in bold. The results for all models can be observed in the Appendix \ref{app:tab:models_by_dataset_perf_models}.}
%   \label{tab:top_dataset_probe_models}
% \end{table}
% Preamble: \usepackage{booktabs,tabularx,multirow}
\begin{table}[t]
\centering
\small
\begin{tabularx}{\linewidth}{>{\raggedright\arraybackslash}p{9mm} X c c c}
\toprule
\textbf{Data} & \textbf{Model} & \textbf{\cvScoreSymb} & \textbf{Layer} & \textbf{Pos} \\
\midrule
\multirow{3}{*}{AMC}
  & \textbf{DeepSeek-Llama-70B}              & \textbf{0.8842} & \textbf{38} & \textbf{-1} \\
  & Qwen2.5-72B-Instruct            & 0.8799 & 61 & -2 \\
  & Qwen2.5-Math-72B                & 0.8780 & 74 & -1 \\
\addlinespace[2pt]
\multirow{3}{*}{CF}
  & \textbf{Llama-3.1-70B-Instruct}          & \textbf{0.8115} & \textbf{42} & \textbf{-1} \\
  & DeepSeek-Llama-70B              & 0.7980 & 41 & -1 \\
  & Llama-3.1-8B-Instruct           & 0.7571 & 16 & -1 \\
\addlinespace[2pt]
\multirow{3}{*}{G8K}
  & \textbf{Llama-3.1-70B-Instruct}          & \textbf{0.5799} & \textbf{35} & \textbf{-1} \\
  & DeepSeek-Llama-70B              & 0.5639 & 38 & -3 \\
  & OpenMath2-Llama3.1-8B           & 0.5516 & 15 & -5 \\
\bottomrule
\end{tabularx}
\caption{Top-3 probe performance scores per dataset. Probe Performance is the average Spearman Rank correlation over 5 folds.Data abbreviations: AMC = E2H-AMC, CF = E2H-Codeforces, G8K = E2H-GSM8K. The results for all models can be observed in the Appendix \ref{app:tab:all_probe_results}}
\label{tab:top_dataset_probe_models}
\end{table}

As shown in Table \ref{tab:top_dataset_probe_models}, the highest probe performance is achieved with activations extracted from the largest models in our suite, with E2H-AMC reaching \cvScoreSymb$=0.88$, followed by E2H-Codeforces (\cvScoreSymb$=0.81$) and E2H-GSM8K (\cvScoreSymb$=0.58$). Notably, all top-performing probes are trained using activations from Llama-based models, with the best AMC probe using the R1-distilled variant. 

% The stark performance gap ($\approx0.3$) between E2H-AMC, E2H-Codeforces, and E2H-GSM8K indicates that human difficulty is better encoded within the activations of LLMs. A plausible explanation for this could be that the difficulty scores generated via various LLM attempts at problems are too noisy, given the varying performance of LLMs on math problems. However, contrasting the E2H-AMC and E2H-GSM8K probes indicates that LLMs know how difficult a math problem is according to human beings whilst struggling to predict how difficult multiple other LLM's might find a problem. This reconciles with previous work that demonstrates that LLM's are bad at predicting how difficult a problem is when prompted. Our contribution is that we find that this is also the case internally for both math and coding problems. 
% As shown in Table~\ref{tab:top_dataset_probe_models}, the highest probe performance is achieved using activations from the largest models in our suite, with E2H-AMC reaching $\cvScoreSymb=0.88$, followed by E2H-Codeforces ($\cvScoreSymb=0.81$) and E2H-GSM8K ($\cvScoreSymb=0.58$). Notably, all top-performing probes use activations from Llama-based models, with the best AMC probe leveraging the R1-distilled variant.

The substantial performance gap ($\approx0.3$) between E2H-AMC and E2H-GSM8K reveals that human difficulty ratings are better encoded in model activations than LLM-derived difficulty estimates. One potential explanation is that difficulty scores derived from LLM problem-solving attempts are inherently noisy due to the high variance in LLM performance on mathematical problems. However, the contrast between E2H-AMC and E2H-GSM8K probes suggests a more nuanced picture: models encode representations of human-perceived difficulty while struggling to predict how difficult other LLMs will find the same problems. This aligns with prior work showing that LLMs perform poorly when explicitly prompted to predict problem difficulty. Our contribution extends this finding by demonstrating that this limitation persists at the representational level for both mathematical and coding tasks.

\begin{figure*}[t]
    \centering
    \includegraphics[width=\linewidth]{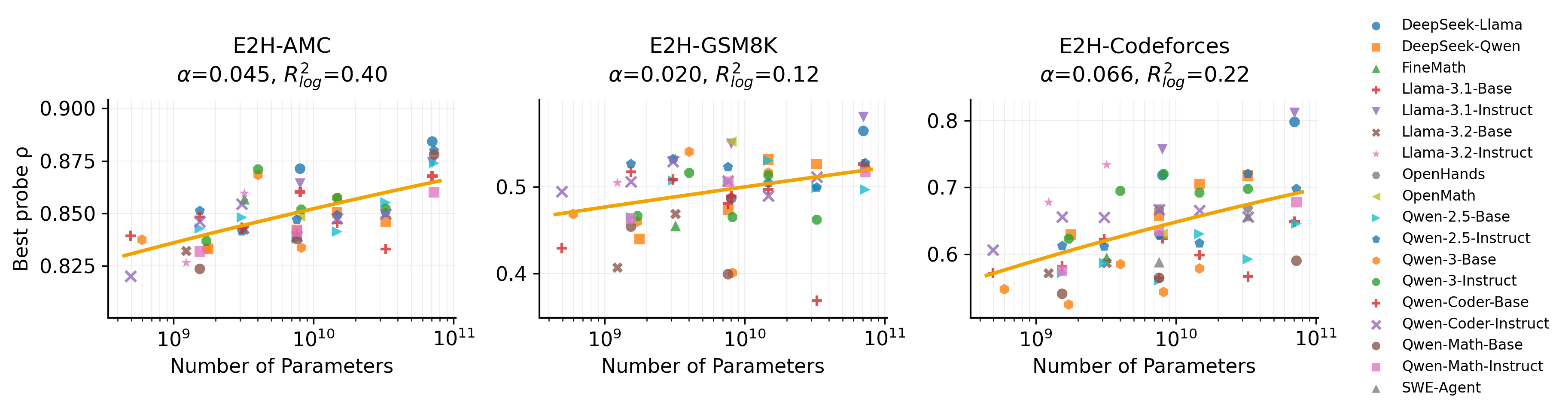}
    % \caption{Human difficulty is more linearly encoded in LLM activations than LLM labels across model architectures. Additionally, difficulty labels exhibit a clear scaling trend with an increase in the number of model parameters. This trend persists across all datasets, although weaker in E2H-GSM8K}
    \caption{\textbf{Probe performance scales with model size following a power law}. The scatterplot shows the best performing probes for various models on each of the three chosen datasets. Each point represents the best-performing linear probe for a given model. We fit a power law in log space on the gap to perfect probe performance (1 - performance) as a function of model size. From this, we observe a scaling trend that persists across all datasets, but is strongest in E2H subsets with human labels (AMC and Codeforces), rather than GSM8K with LLM labels.}
    
    \label{fig:full_cross_model_comparison}
\end{figure*}

 % as perceived by humans is more consistently represented in LLM internal states than difficulty estimated by models themselves.
    % }

% \paragraph{Difficulty representations scale with model size} Figure~\ref{fig:full_cross_model_comparison} shows that probe performance ($\cvScoreSymb$) follows a power law with respect to model parameters across all datasets. E2H-AMC exhibits moderate scaling ($\alpha=0.045$), with probe scores ranging from 0.82 to 0.89. E2H-Codeforces demonstrates the strongest scaling ($\alpha=0.066$), indicating that coding difficulty representations benefit most from increased model capacity. In contrast, E2H-GSM8K shows the weakest scaling ($\alpha=0.020$), suggesting that scale alone provides limited improvement for this dataset.

% The clearer scaling trend observed for human difficulty labels (E2H-AMC and E2H-Codeforces) compared to LLM-derived labels (E2H-GSM8K tasks) indicates that human-perceived difficulty is more consistently encoded in model representations than model-estimated difficulty. The weak scaling for GSM8K likely reflects task-specific bottlenecks in individual models that prevent larger architectures from forming better difficulty representations, such as Llama-3.1-32B-Base. The higher  $R^2_{log}$ of 0.40 in the E2H-AMC dataset might explain why the scaling trend is more consistent than in the case of E2H-GSM8K where $R^2_{log}=0.12$.

\paragraph{Difficulty representations scale with model size}Figure~\ref{fig:full_cross_model_comparison} demonstrates that probe performance (\cvScoreSymb) follows a power law with model parameters across all three datasets. The scaling strength varies by dataset: E2H-Codeforces shows the strongest scaling ($\alpha=0.066$), suggesting that larger models develop better representations of coding difficulty. E2H-AMC exhibits moderate scaling ($\alpha=0.045$), with probe scores improving from 0.82 to 0.89 as model size increases. E2H-GSM8K shows the weakest scaling ($\alpha=0.020$), indicating that model size alone provides limited benefit for LLM-derived difficulty scores.

The scaling patterns reveal an important distinction between human-labeled and LLM-labeled difficulty. Datasets with human difficulty labels (E2H-AMC and E2H-Codeforces) exhibit clearer scaling trends than those with LLM-derived labels (E2H-GSM8K), suggesting that models more consistently encode human-perceived difficulty in their representations. The weak scaling for GSM8K likely stems from task-specific bottlenecks in individual models, as seen in Llama-3.1-32B-Base in Figure ~\ref{fig:full_cross_model_comparison}, that prevent larger architectures from forming better difficulty representations. This interpretation is supported by the fit quality: E2H-AMC achieves $R^2_{log}$=0.40, indicating a strong power law relationship for human-rated math problems, while E2H-GSM8K's $R^2_{log}$=0.12 reflects a much weaker and noisier trend. This tells us that human difficulty in math problems is robustly encoded amongst a suite of models as compared to LLM-sourced difficulty labels for math questions..

\paragraph{The Optimal Probing Position Varies by Task Domain}
% \begin{figure}[t]
%     \centering
%     \includegraphics[width=\linewidth]{figs/position_heatmap_excluding_DeepSeek-Qwen.png}
%     \caption{Distribution of optimal token positions for difficulty-predicting probes across datasets. For each model, we identify the post-instruction token position yielding the highest probe performance and count how many models achieve their best results at each position. Positions are indexed from the sequence end ($-1$ = final token). DeepSeek-Qwen is excluded due to its extended post-instruction sequences; the complete figure is available in Appendix~\ref{fig:full_post_token_analysis}.}
%     \label{fig:position_heatmap}
% \end{figure}
\begin{figure}[t]
    \centering
    \includegraphics[width=\columnwidth]{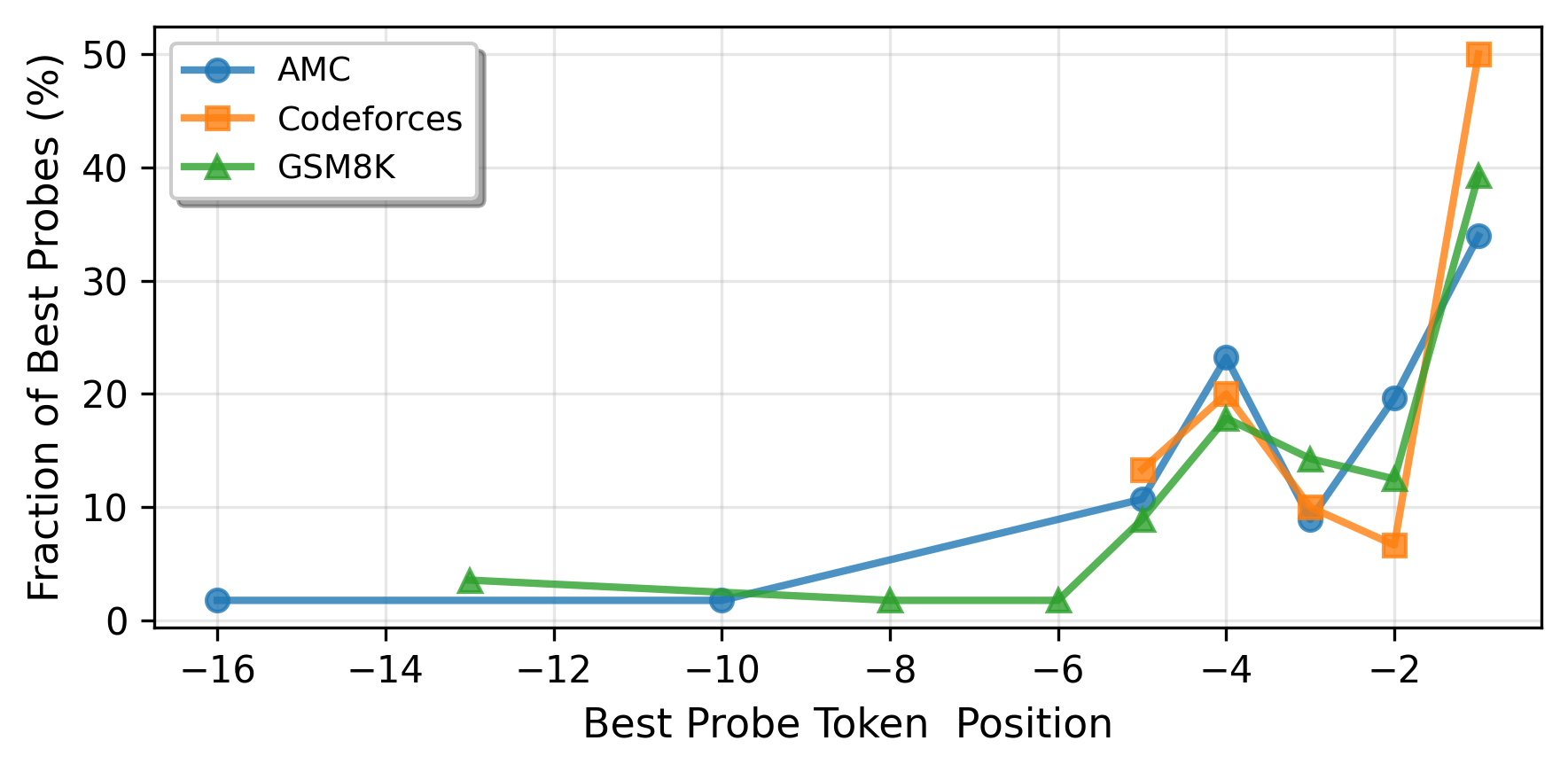}
    \caption{\textbf{Optimal probe positions vary by domain.} Percentage of best-performing probes at each token position (negative indices from end of prompt). 
Codeforces shows strong preference for the final token (50\% of probes), while mathematical datasets (AMC, GSM8K) distribute more evenly across recent tokens, suggesting that difficulty signals are encoded differently across domains.}
    \label{fig:full_post_token_analysis}
\end{figure}

The best performing probes train on activations from middle layers at the final token position. However, Figure~\ref{fig:full_post_token_analysis} reveals that while the final token generally yields optimal probes across datasets, certain positions (-4 and -2) show strong performance for specific models, indicating that the final token is not universally superior.

This pattern is most evident when comparing Codeforces with the mathematical datasets (AMC and GSM8K). For Codeforces, 50\% of the best probes are trained at the last token, with other probe positions within the last five tokens across all models. In contrast, the mathematical datasets range from -16 to -1 despite having 4 fewer models to train probes with. This strongly suggests that coding difficulty is more strongly encoded near the prompt's end than in other task domains.

% \paragraph{Steering toward easier difficulty induces tool use reasoning and boosts MATH performance.} 
\begin{figure}[t]
    \centering
    \includegraphics[width=\columnwidth]{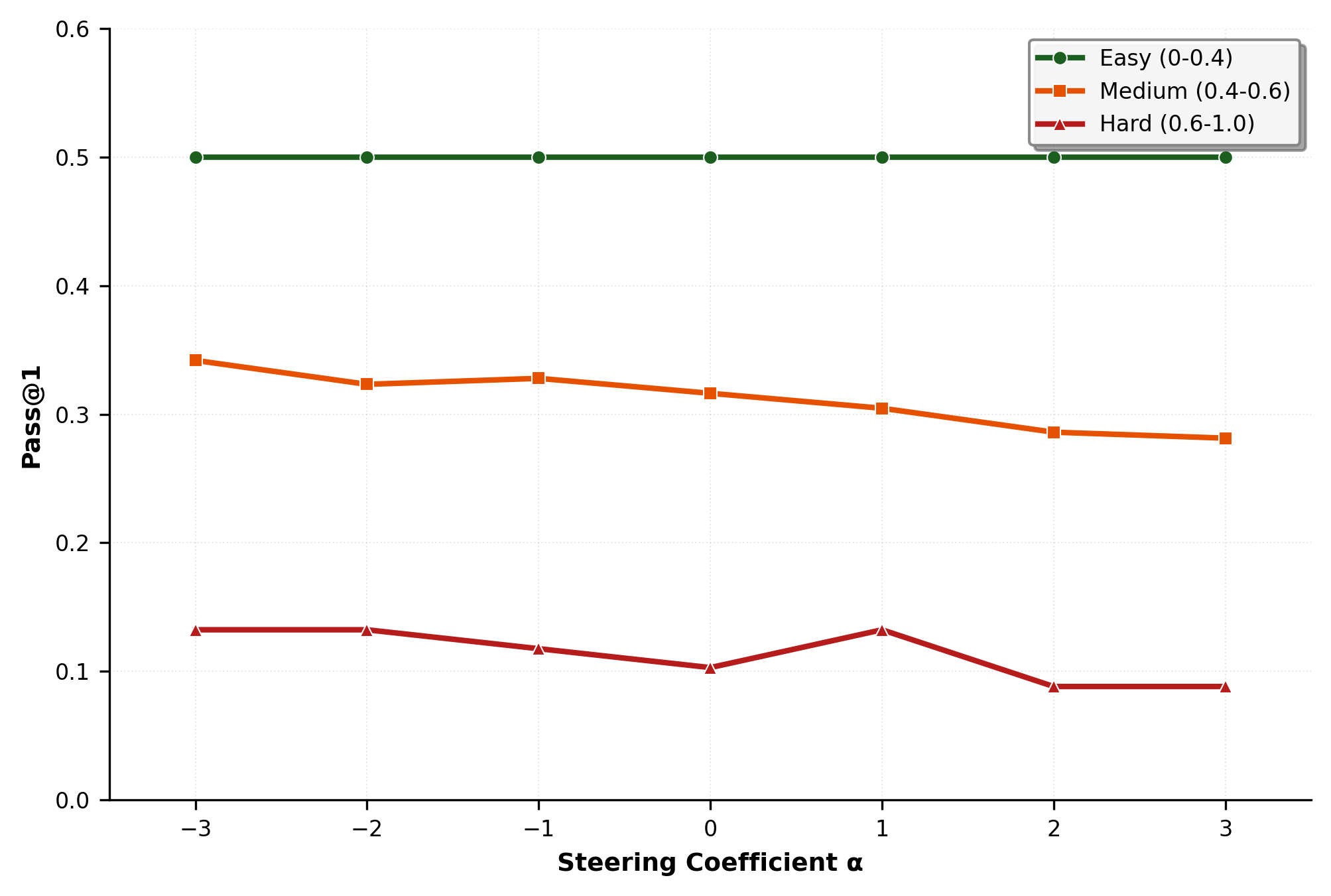}
    \caption{\textbf{Steering towards "easy" improves mathematical reasoning performance.} Pass@1 accuracy on MATH500 for Qwen2.5-Math-1.5B across steering coefficients $\alpha \in [-3, 3]$. Negative coefficients (easier direction) improve performance across all difficulty bins, while positive coefficients (harder direction) degrade performance, particularly on medium and hard problems.}
    \label{fig:steered_predicted_performance}
\end{figure}

\paragraph{Linear probes can steer models towards longer generations and induce tool reasoning}
Given the strong performance of our linear probes on the E2H-AMC subset, we assume a linear "easy-hard" axis exists within LLM representations in math settings. Following \citet{li_inference-time_2024}, we use the fitted regression coefficients from our trained probes to steer generation during inference. We explore steering with difficulty vectors on Qwen2.5-Math-1.5B's Pass@1 performance on the MATH500 benchmark \cite{hendrycks_measuring_2021}, using the steering coefficient grid defined in Section \ref{subsec:Probe_Setup}.

In Figure \ref{fig:steered_predicted_performance}, we visualize Qwen2.5-Math-1.5B's performance on MATH500 for each steering coefficient $\alpha$. Steering towards the "easy" (negative) direction improves performance across all predicted difficulty bins, particularly at $\alpha=-3$. Surprisingly, steering towards the "hard" direction reduces performance on medium and hard questions, with performance gains on hard questions only at $\alpha=1$.

As shown in Figure \ref{fig:qual-steer-card}, at $\alpha=0$ or $\alpha=+3$, the model hallucinates, fails to generate an answer, and reaches maximum generation length. In contrast, at $\alpha=-3$, the model generates code before and after producing the correct answer. This pattern holds across most questions: steering with $\alpha=-3$ reduces generated text length and increases code generation during inference. We attach more of these results in the Appendix \ref{app:fig:steering_len_hist} and \ref{app:fig:steering_py_hist}.

\begin{figure}[t]
    \centering
    \includegraphics[width=\columnwidth]{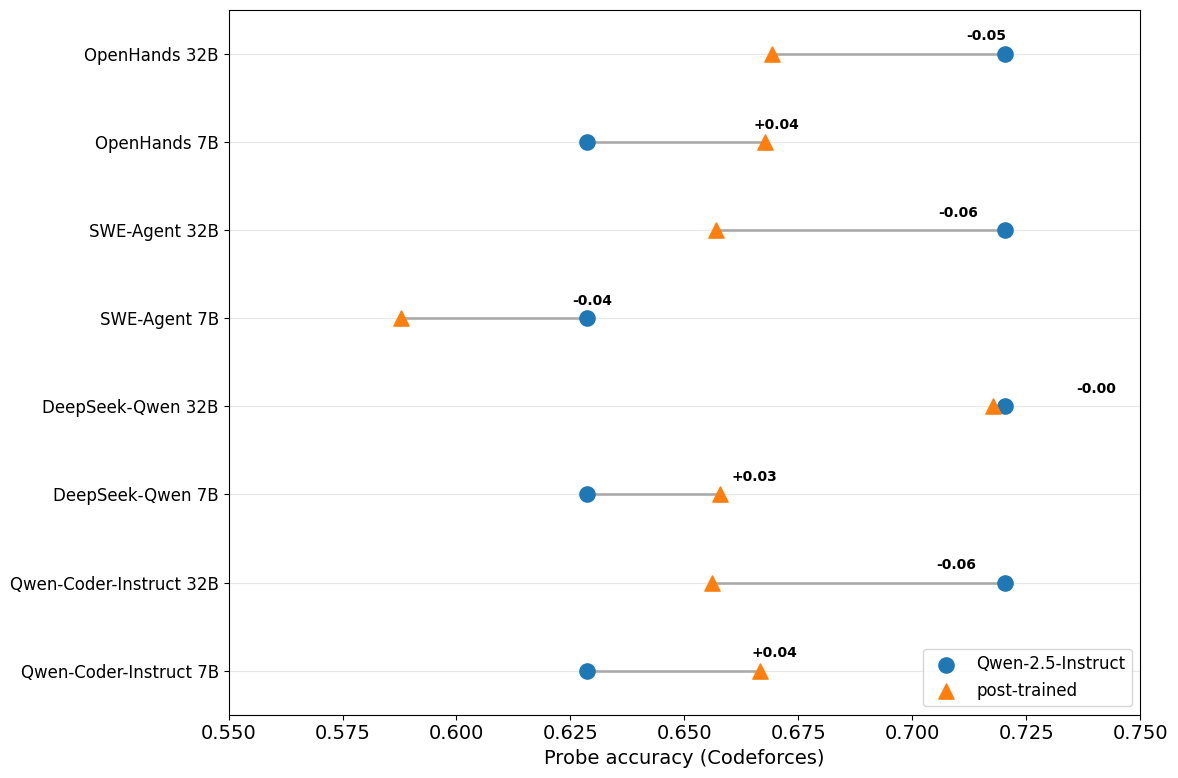}
     \caption{\textbf{Specialised fine-tuning shows diminishing returns for larger models.} Probe accuracy on Codeforces for Qwen-2.5-Instruct and its specialised variants. Values show accuracy change after task-specific fine-tuning: 7B models improve modestly, while most 32B models degrade slightly.}
    \label{fig:post_train_comparison}
\end{figure}

\paragraph{Does specialised post-training improve probe performance? }
Specialised fine-tuning beyond instruction tuning shows mixed results. Figure~\ref{fig:post_train_comparison} compares Qwen-2.5-Instruct models (which have undergone instruction tuning) against specialised variants that receive additional task-specific fine-tuning. While some 7B models benefit modestly from the additional specialisation (OpenHands-7B: +0.04, DeepSeek-Qwen-7B: +0.03, Qwen-Coder-Instruct-7B: +0.04), most 32B models show slight degradation (OpenHands-32B: -0.05, SWE-Agent-32B and Qwen-Coder-Instruct-32B: -0.06 each). This suggests that instruction tuning already enables models to form robust difficulty representations, particularly in larger models. Additional task-specific fine-tuning may help smaller models refine these representations, but risks disrupting the general difficulty understanding acquired during instruction tuning in larger models.

\subsection{Tracking Difficulty Representations during GRPO}
If our probes have identified a linear "easy-hard" direction, the next natural question is what happens to this direction during post-training. To operationalise this, we investigate how GRPO for math reasoning affects the linear representations of difficulty and whether this evolution depends on the source of difficulty labels (human vs. LLM annotations).

\paragraph{Experimental Design.} We train probes on model checkpoints throughout GRPO training using both E2H-AMC (human-labelled) and E2H-GSM8K (LLM-labelled) difficulty scores. This allows us to track how difficulty representations evolve and whether human vs. LLM difficulty labels show different adaptation patterns. Due to computational constraints, we focus on training probes for each checkpoint of Qwen2.5-Math-1.5B. We perform GRPO with Llama-3.2-3B-Instruct but place these results in the Appendix \ref{app:fig:grpo_training_curves}.

\paragraph{GRPO Configuration} Unless otherwise noted, we follow the Dr.\,GRPO recipe \cite{liu_understanding_2025}.We use the MATH training split with 12,000 examples and filtered it to problems with an annotated difficulty $\ge3$, yielding 8,890 questions. Our changes are that training utilises the Verl library on a single A100-80GB GPU with a batch size of 256 and 64 gradient accumulation steps. We increase the learning rate from $1 \times 10^{-6}$ to $1 \times 10^{-5}$ based on preliminary experiments, with all other hyperparameters remaining the same as in Dr GRPO. We save a checkpoint of the model at each step.

\paragraph{Probe Training.} At each GRPO checkpoint, we extract activations and train linear probes using the same configuration described in Section \ref{subsec:Probe_Setup}. Each checkpoint is treated as an independent model, allowing us to track changes in probes at each training step.

\paragraph{Evaluation.} We track two complementary metrics throughout GRPO training. Model performance is measured with Pass@1 accuracy on MATH500 \cite{hendrycks_measuring_2021}. The answer correctness is verified using \textsc{Math-Verify} \cite{kydlicek_math-verify_2025}, and for the evaluation settings, we use greedy decoding with a max token budget of 3000. Probe quality is as in subsection \ref{subsec:Probe_Setup}. This allows us to observe how difficulty representation evolves alongside mathematical reasoning capability.

\subsubsection{Results}
% \paragraph{GRPO results} From Table \ref{tab:grpo_math500_peak_with_baseline}, we observe that our GRPO configuration beats the Dr.GRPO baselines Oat-Zero-1.5B and Oat-Zero-3B in fewer steps and with a single GPU, effectively setting a new SOTA for the MATH500 benchmark.

\paragraph{GRPO results}
Table~\ref{tab:grpo_math500_peak_with_baseline_and_drgrpo} shows that our Dr.GRPO implementation achieves a higher peak Pass@1 on \textsc{MATH500} than the reported Dr.\,GRPO baseline (Oat-Zero-1.5B: 74.2), using fewer total training steps and with a single GPU. This validates our experimental setup and allows us to study how our probes change during training at each step. Full training curves are provided in Appendix~\ref{app:fig:grpo_training_curves}.

\begin{figure*}[t]
    \centering
    \includegraphics[width=\linewidth]{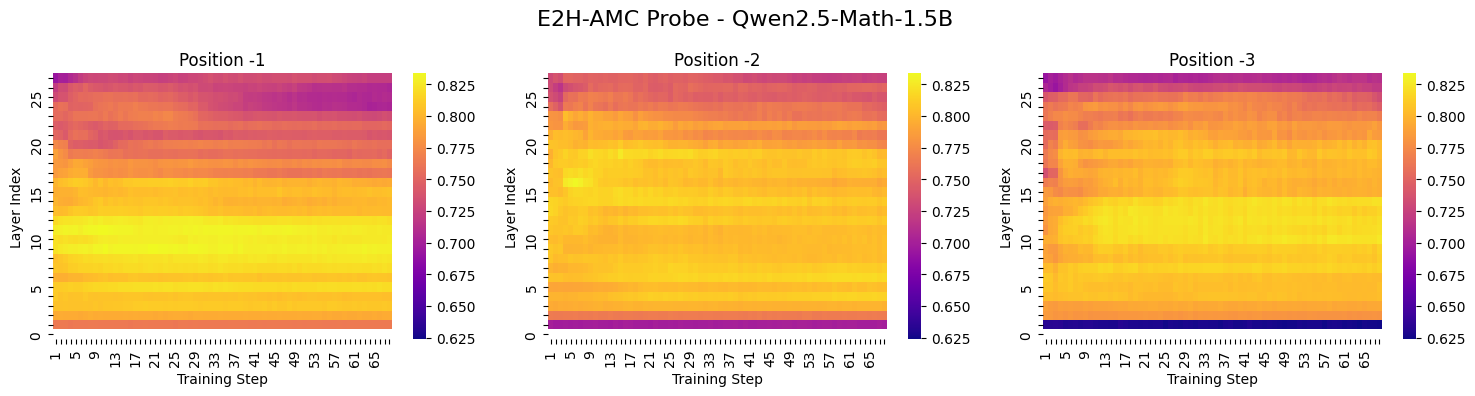}
    \includegraphics[width=\linewidth]{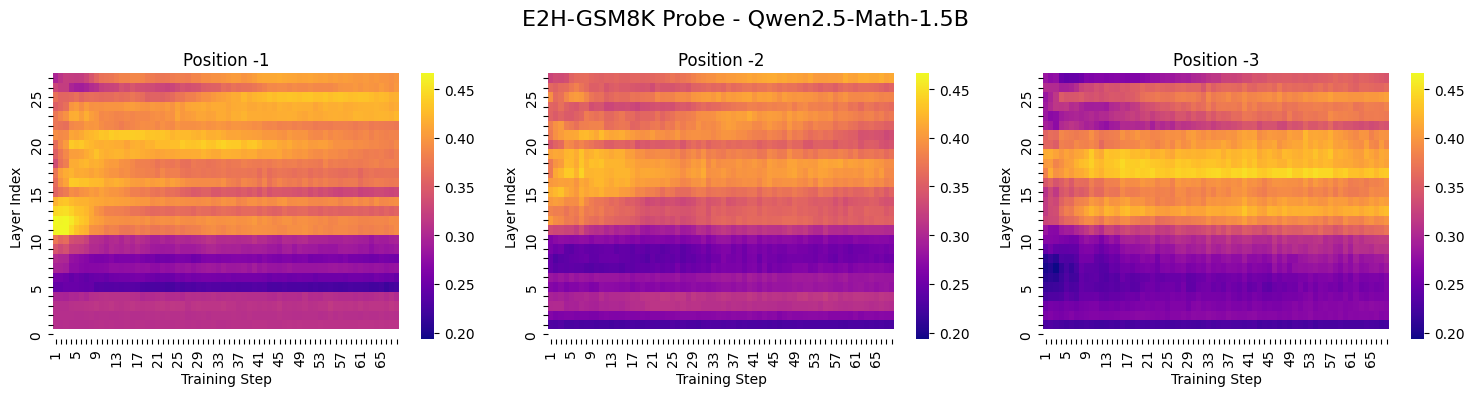}

    \caption{\textbf{LLM-derived difficulty degrades across most layers while human-aligned difficulty remains mostly stable.} Qwen2.5-Math-1.5B linear Probe performance across layers  (y-axis) and training steps (x-axis) for the last three token positions. GSM8K probes (rows 2,4) show pervasive degradation in early and middle layers (up to 50\% decline). AMC probes (Top row) maintain stable or improving performance across all layers and positions.}
    
    \label{fig:grpo_ckpt_heatmap}
\end{figure*}

\begin{table}[t]
\centering
\scriptsize
\setlength{\tabcolsep}{4pt}
\caption{GRPO on \textsc{MATH500}: baseline (step 0) and peak Pass@1. 
Rows under \emph{Dr.GRPO (reported)} are taken from the paper and don't report when the peak step is reached, but give us the total number of training steps when following the training recipe.}
\label{tab:grpo_math500_peak_with_baseline_and_drgrpo}
\begin{tabularx}{\columnwidth}{l c c c}
\toprule
Model & Baseline & Peak Pass@1 & Peak Step \\
\midrule
\multicolumn{4}{l}{\textit{Dr.GRPO (Ours)}}\\
Qwen2.5\textendash Math\textendash 1.5B       & 64.7 & \textbf{76.2} & 43/67 \\
% Llama\textendash 3.2\textendash 3B\textendash Instruct & 46.6 & 53.4 & 54/67 \\
\addlinespace
\multicolumn{4}{l}{\textit{Dr.GRPO \cite{liu_understanding_2025}}}\\
Oat\textendash Zero\textendash 1.5B & 61.8 & 74.2 & NA/11,200 \\
% Oat\textendash Zero\textendash 3B   & 38.8 & 50.0 & NA/11,200 \\
\bottomrule
\end{tabularx}
\end{table}

\paragraph{Human and LLM difficulty representations evolve oppositely during training}

To understand how difficulty representations change during GRPO, we track probe performance across training checkpoints. Figure~\ref{fig:grpo_ckpt_heatmap} reveals contrasting patterns: human-aligned difficulty probes (E2H-AMC, top row) remain stable or improve across training steps, particularly in middle layers. In contrast, LLM-derived difficulty probes (E2H-GSM8K, bottom row) show pervasive degradation in early and middle layers, with performance declining by up to 50\%. This layer-wide degradation suggests that LLM-derived difficulty is a noisy signal that GRPO systematically overwrites.

\begin{figure*}[t]
    \centering
    \includegraphics[width=\linewidth]{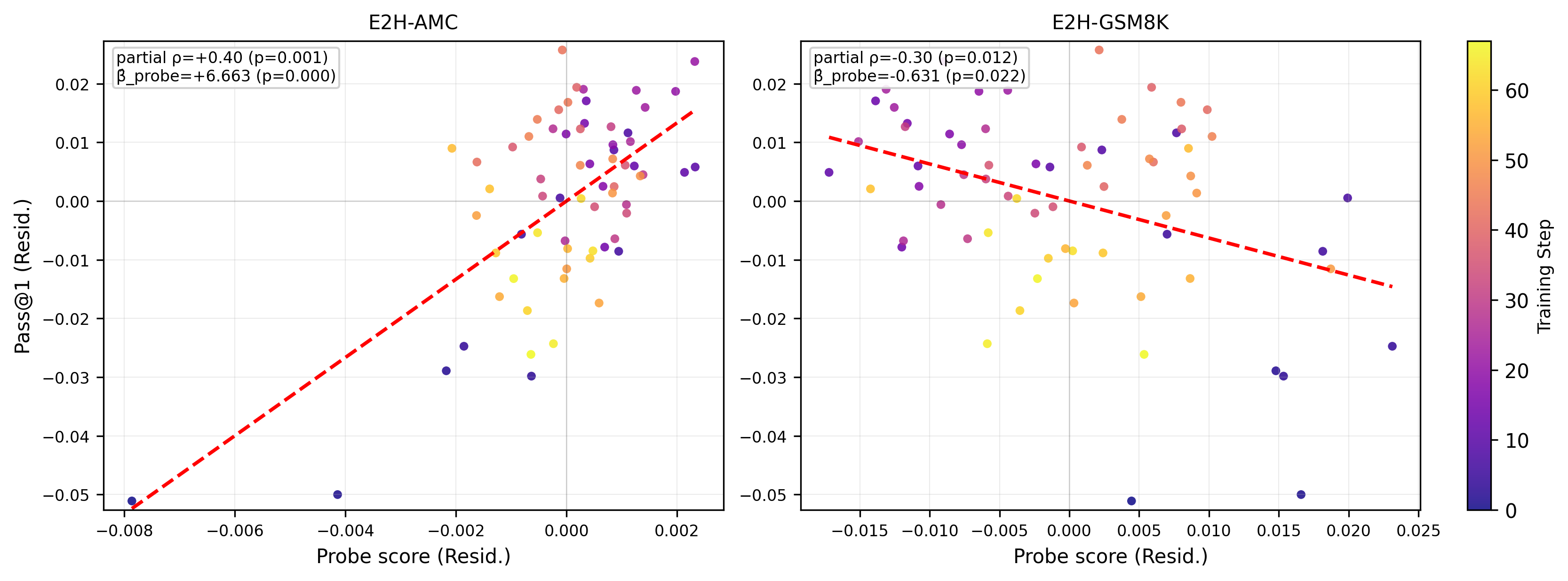}
    \caption{\textbf{Human-difficulty encoding correlates positively with performance; LLM-difficulty encoding correlates negatively.} Each point represents a training checkpoint (color indicates step) for Qwen2.5-Math-1.5B. Axes show residuals after removing the linear effect of training step, isolating the relationship between probe performance (x-axis) and test Pass@1 (y-axis) independent of overall training progress. Human-labeled difficulty (AMC, left) shows strong positive correlation ($\beta$=+6.66, p<0.001), while LLM-derived difficulty (GSM8K, right) shows negative correlation ($\beta$=-0.63, p=0.022).}
    \label{fig:scatter_probe_x_grpo_residuals}
\end{figure*}

To test whether these representational changes relate to task performance, we examine the correlation between probe strength and Pass@1 accuracy after controlling for overall training progress. Figure~\ref{fig:scatter_probe_x_grpo_residuals} plots residuals from regressing both variables on training step, revealing the probe-performance relationship independent of general improvement. For human-labeled difficulty (AMC), stronger probe performance strongly predicts higher test accuracy ($\beta$=+6.66, p<0.001). For LLM-derived difficulty (GSM8K), the relationship inverts: stronger probe performance predicts lower test accuracy ($\beta$=-0.63, p=0.022). This confirms that human difficulty representations strengthen alongside capabilities, while LLM-derived representations degrade as models improve—consistent with the layer-wise degradation in Figure~\ref{fig:grpo_ckpt_heatmap}.

% Figure~\ref{fig:scatter_probe_x_grpo_residuals} demonstrates that optimal probe performance diverges by difficulty source: human-aligned probes improve or remain stable ($\rho$+0.95, +0.60), while LLM-derived probes degrade ($\rho$=-0.98, -0.95). To  understand whether this reflects genuine representational changes or selection artifacts, we examine probe performance across all layers and 
% training steps (Figure~\ref{fig:grpo_ckpt_heatmap}).

% For GSM8K, degradation is pervasive: early and middle layers decline by up to 50\%, making it impossible to maintain performance through alternative layer selection. For AMC, probes either improve (Qwen, middle layers) or remain stable  (Llama), confirming that best-probe improvements reflect genuine changes in difficulty encoding rather than selection effects.

\section{Limitations}
This work has several key limitations. First, we focus exclusively on coding and math tasks from E2H, omitting three other subsets. We selected these domains to compare difficulty representations from LLM versus human judgments, with Codeforces as an out-of-domain baseline. Computational constraints prevented extensive experiments across all E2H subsets and GRPO training on larger models. Similarly, memory constraints from training probes and saving checkpoints prevented tracking multiple GRPO seeds.

Our steering experiments focus on Qwen2.5-Math-1.5B, which motivated our RL experiments. However, we recognise that the observed steering effects may be model-specific and broader cross-model studies are central to characterising the full effect of steering along the difficulty direction.

Future work should investigate whether these probes generalise to other agentic tasks requiring coding and reasoning beyond Codeforces and explore how difficulty representations evolve during inference and multi-turn conversations.

\section{Conclusion}
We demonstrate that difficulty exists as a linear direction in LLM activations that aligns more strongly with human judgment than LLM-derived estimates. Across 60 models, human-labelled difficulty shows robust encoding and clear scaling laws, while LLM-derived difficulty exhibits weak encoding and poor scaling. This reveals a critical asymmetry: models maintain strong implicit representations of human-perceived difficulty but struggle to articulate it explicitly, suggesting internal representations capture signals inaccessible through generation.

Steering experiments on Qwen2.5-Math-1.5B show functional consequences: steering toward "easier" ($\alpha$ = -3, -2, -1) improves MATH500 accuracy while reducing length and hallucination; steering toward "harder" degrades accuracy and increases both. Easier steering also increases code generation, mirroring patterns from GRPO training, suggesting shared mechanisms.

During GRPO training, we also observed that human-aligned difficulty probes strengthened with task performance, while LLM-aligned difficulty probes declined. Combined with steering results, this suggests GRPO refines difficulty representations in ways that align with effective reasoning strategies such as tool use. Collectively, these gains indicate that refined difficulty representation may be functionally important for math reasoning and potentially other tasks.

Future work should investigate why easier steering enhances reasoning, whether these patterns generalise to other domains, and whether adaptive steering during inference time could boost performance. 

\FloatBarrier

% \bibliography{custom}

\clearpage
\appendix
\section{Appendix}
\label{sec:appendix}
\FloatBarrier

\begin{table}[t]
\centering
\scriptsize
\caption{Model families with sizes and variant counts}
\label{app:tab:models_list}
\begin{tabular}{lp{3cm}c}
\toprule
Model Family & Sizes (B) & Variants \\
\midrule
\multicolumn{3}{l}{\textit{Base Models}} \\
Qwen-2.5-Base & 1.5, 3, 7, 14, 32, 72 & 6 \\
Qwen-3-Base & 0.6, 1.7, 4, 8, 14, 32 & 6 \\
Qwen-Coder-Base & 0.5, 1.5, 3, 7, 14, 32 & 6 \\
Qwen-Math-Base & 1.5, 7, 72 & 3 \\
Llama-3.2-Base & 1, 3 & 2 \\
Llama-3.1-Base & 8, 70 & 2 \\
\addlinespace
\multicolumn{3}{l}{\textit{Instruct Models}} \\
Qwen-2.5-Instruct & 1.5, 3, 7, 14, 32, 72 & 6 \\
Qwen-Coder-Instruct & 0.5, 1.5, 3, 7, 14, 32 & 6 \\
Qwen-3-Instruct & 1.7, 4, 8, 14, 32 & 5 \\
Qwen-Math-Instruct & 1.5, 7, 72 & 3 \\
Llama-3.1-Instruct & 8, 70 & 2 \\
Llama-3.2-Instruct & 1, 3 & 2 \\
\addlinespace
\multicolumn{3}{l}{\textit{Specialized Models}} \\
DeepSeek-Qwen & 1.5, 7, 14, 32 & 4 \\
DeepSeek-Llama & 8, 70 & 2 \\
OpenHands & 7, 32 & 2 \\
SWE-Agent & 7, 32 & 2 \\
FineMath & 3 & 1 \\
OpenMath & 8 & 1 \\
\bottomrule
\end{tabular}
\end{table}
% \columnbreak

\subsection{Probe Results for all Models}

\begin{tcolorbox}[appbox,title={E2H-AMC}]
\RaggedRight
Please take your time to thoroughly analyze and solve the following math competition problem step by step. Your approach should be detailed, ensuring that each step of your reasoning is clearly explained to minimize errors and maximize understanding.

[PROBLEM\_START] \{problem\} [PROBLEM\_END]

While solving, consider all possible scenarios and subtleties involved in the problem. Each step should build upon the previous one logically, leading to a cohesive solution.

Once you arrive at the solution, please present the final answer enclosed in ` ' `. Ensure the answer is displayed using appropriate \LaTeX{} formatting to maintain mathematical precision and clarity.
\end{tcolorbox}

\begin{tcolorbox}[appbox,title={E2H-Codeforces}]
\RaggedRight
Please generate executable Python 3.10 code that directly solves the problem described below. The code should be ready to run without any modifications or additional comments. It must strictly follow Python 3.10 syntax and be formatted correctly for direct execution. Do not include explanations or comments within the code.

[PROBLEM\_MAIN\_START] \{problem\_main\} [PROBLEM\_MAIN\_END]

[PROBLEM\_NOTE\_START] \{problem\_note\} [PROBLEM\_NOTE\_END]

[INPUT\_SPEC\_START] \{input\_spec\} [INPUT\_SPEC\_END]

[OUTPUT\_SPEC\_START] \{output\_spec\} [OUTPUT\_SPEC\_END]

[SAMPLE\_INPUTS\_START] \{sample\_inputs\} [SAMPLE\_INPUTS\_END]

[SAMPLE\_OUTPUTS\_START] \{sample\_outputs\} [SAMPLE\_OUTPUTS\_END]

\begin{tightenum}
  \item Include any required imports at the top of the script.
  \item Use \verb|sys.stdin.readline()| instead of \verb|input()|.
  \item Begin with \verb|"""python| and end with \verb|"""|. The content between must be valid Python~3.10 and directly savable as a \texttt{.py} file.
\end{tightenum}
\end{tcolorbox}
\FloatBarrier

% --- Auto-generated single-column tables per dataset (descending, bold best) ---
% --- Auto-generated wide model x dataset matrix ---
\begin{table*}[tbp]
\centering
\small
\caption{Linear probe results for all models (higher is better). Best per dataset in \textbf{bold}.}
\label{app:tab:all_probe_results}
\begin{adjustbox}{max width=\textwidth}
\begin{tabular}{lccc}
\toprule
\textbf{Model} & \textbf{E2H-AMC} & \textbf{E2H-GSM8K} & \textbf{E2H-Codeforces} \\
\midrule
Llama-3.1-70B-Instruct & 0.874 & \textbf{0.580} & \textbf{0.811} \\
DeepSeek-R1-Distill-Llama-70B & \textbf{0.884} & 0.564 & 0.798 \\
Llama-3.1-8B-Instruct & 0.864 & 0.549 & 0.757 \\
Llama-3.2-3B-Instruct & 0.860 & 0.533 & 0.734 \\
Qwen2.5-72B-Instruct & 0.880 & 0.527 & 0.698 \\
DeepSeek-R1-Distill-Qwen-32B & 0.846 & 0.526 & 0.718 \\
DeepSeek-R1-Distill-Qwen-14B & 0.851 & 0.531 & 0.705 \\
Qwen3-4B & 0.871 & 0.516 & 0.695 \\
DeepSeek-R1-Distill-Llama-8B & 0.871 & 0.487 & 0.718 \\
Qwen2.5-32B-Instruct & 0.850 & 0.499 & 0.720 \\
Qwen3-14B & 0.857 & 0.513 & 0.692 \\
Qwen2.5-Math-72B-Instruct & 0.860 & 0.517 & 0.678 \\
Llama-3.1-70B & 0.868 & 0.526 & 0.649 \\
Qwen2.5-Coder-3B-Instruct & 0.854 & 0.529 & 0.655 \\
Qwen3-8B & 0.852 & 0.465 & 0.720 \\
OpenMath2-Llama3.1-8B & 0.849 & 0.552 & 0.630 \\
Qwen2.5-Coder-32B-Instruct & 0.850 & 0.511 & 0.656 \\
Qwen2.5-72B & 0.874 & 0.497 & 0.646 \\
Qwen2.5-Coder-7B-Instruct & 0.839 & 0.507 & 0.667 \\
Qwen3-32B & 0.852 & 0.462 & 0.698 \\
Llama-3.2-1B-Instruct & 0.826 & 0.504 & 0.677 \\
Qwen2.5-Coder-1.5B-Instruct & 0.846 & 0.506 & 0.656 \\
openhands-lm-32b-v0.1 & -- & -- & 0.669 \\
openhands-lm-7b-v0.1 & -- & -- & 0.668 \\
Qwen2.5-Coder-14B-Instruct & 0.847 & 0.489 & 0.665 \\
Qwen2.5-14B & 0.841 & 0.530 & 0.630 \\
Qwen2.5-7B-Instruct & 0.847 & 0.522 & 0.629 \\
Qwen3-4B-Base & 0.868 & 0.540 & 0.585 \\
Qwen2.5-Math-72B & 0.878 & 0.522 & 0.590 \\
Qwen2.5-1.5B-Instruct & 0.851 & 0.526 & 0.612 \\
Qwen2.5-3B-Instruct & 0.841 & 0.532 & 0.612 \\
Qwen2.5-Math-7B-Instruct & 0.842 & 0.506 & 0.635 \\
Qwen2.5-Coder-3B & 0.843 & 0.508 & 0.622 \\
DeepSeek-R1-Distill-Qwen-7B & 0.842 & 0.473 & 0.658 \\
Llama-3.1-8B & 0.860 & 0.489 & 0.624 \\
SWE-agent-LM-32B & -- & -- & 0.657 \\
Qwen2.5-14B-Instruct & 0.849 & 0.504 & 0.616 \\
Qwen3-14B-Base & 0.857 & 0.516 & 0.578 \\
Qwen2.5-Coder-1.5B & 0.848 & 0.517 & 0.582 \\
Qwen2.5-32B & 0.855 & 0.499 & 0.592 \\
Qwen2.5-3B & 0.848 & 0.507 & 0.587 \\
Qwen2.5-Coder-14B & 0.845 & 0.497 & 0.599 \\
Qwen3-1.7B & 0.837 & 0.466 & 0.623 \\
Qwen2.5-Coder-0.5B-Instruct & 0.820 & 0.494 & 0.606 \\
FineMath-Llama-3B & 0.857 & 0.455 & 0.594 \\
Qwen2.5-7B & 0.838 & 0.506 & 0.561 \\
DeepSeek-R1-Distill-Qwen-1.5B & 0.833 & 0.440 & 0.629 \\
Llama-3.2-3B & 0.842 & 0.469 & 0.587 \\
Qwen2.5-Coder-7B & 0.840 & 0.480 & 0.565 \\
Qwen2.5-1.5B & 0.843 & 0.465 & 0.572 \\
Qwen2.5-Math-1.5B-Instruct & 0.832 & 0.464 & 0.575 \\
Qwen3-0.6B-Base & 0.837 & 0.469 & 0.547 \\
Qwen2.5-Coder-0.5B & 0.839 & 0.429 & 0.571 \\
Qwen2.5-Math-1.5B & 0.824 & 0.454 & 0.541 \\
Qwen3-1.7B-Base & 0.833 & 0.460 & 0.524 \\
Llama-3.2-1B & 0.832 & 0.407 & 0.571 \\
Qwen2.5-Math-7B & 0.838 & 0.399 & 0.564 \\
Qwen3-8B-Base & 0.834 & 0.401 & 0.543 \\
Qwen2.5-Coder-32B & 0.833 & 0.369 & 0.566 \\
SWE-agent-LM-7B & -- & -- & 0.588 \\
% gemma-3-1b-it & -- & -- & 0.495 \\
\bottomrule
\end{tabular}
\end{adjustbox}
\end{table*}

% \subsection{GRPO Exp}
\begin{figure}[t]
    \centering
    \includegraphics[width=\columnwidth]{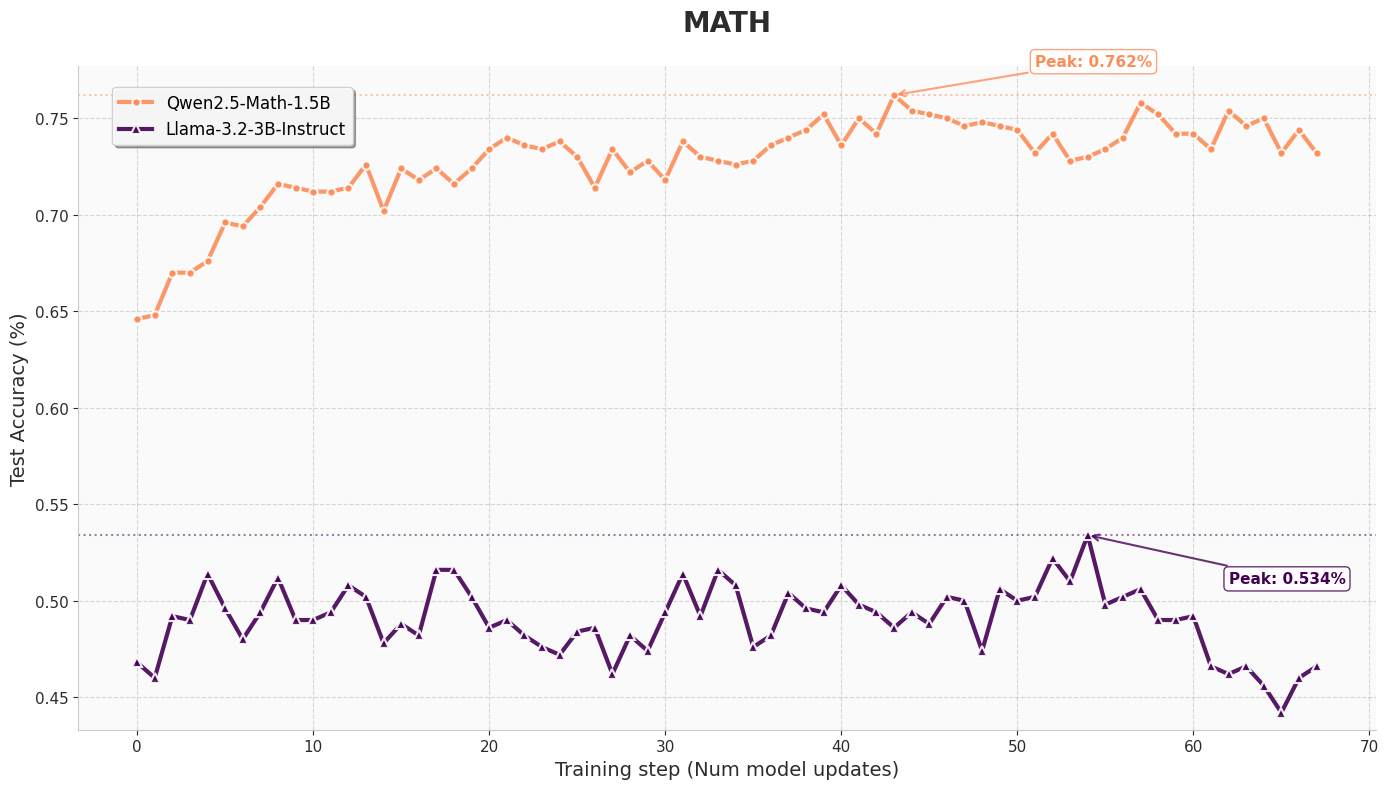}
    \caption{GRPO Training Curves for Qwen and Llama}
    \label{app:fig:grpo_training_curves}
\end{figure}

\begin{figure}[t]
    \centering
    \includegraphics[width=\columnwidth]{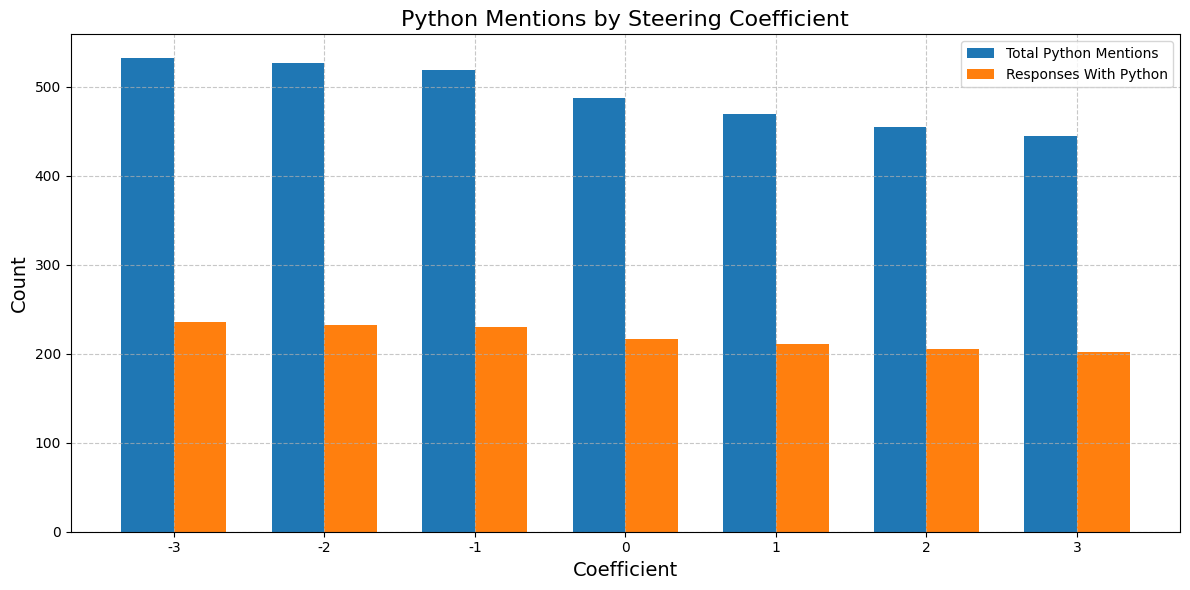}
    \caption{Steering towards the easy direction increases the amount of Python code generated during inference}
    \label{app:fig:steering_py_hist}
\end{figure}

\begin{figure}[t]
    \centering
    \includegraphics[width=\columnwidth]{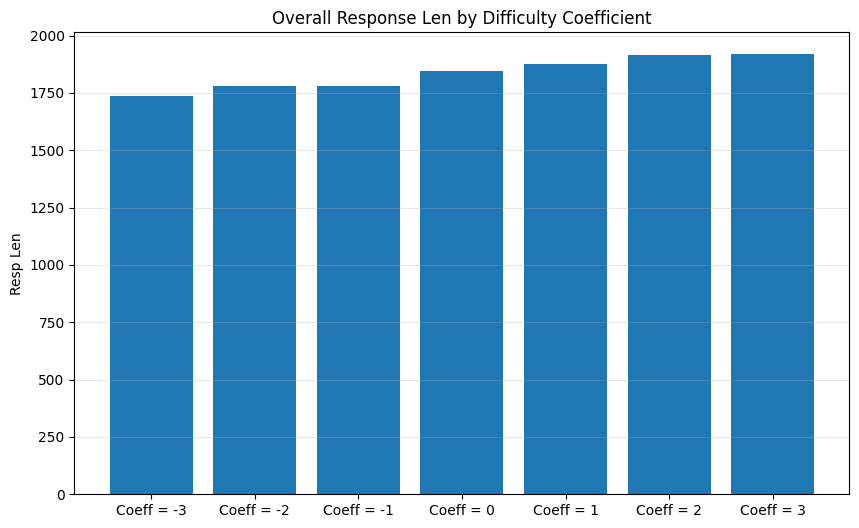}
    \caption{Steering towards the hard direction increases the response length.}
    \label{app:fig:steering_len_hist}
\end{figure}

\end{document}